\documentclass{article}

\usepackage{PRIMEarxiv}

\usepackage[utf8]{inputenc} % allow utf-8 input
\usepackage[T1]{fontenc}    % use 8-bit T1 fonts
\usepackage{hyperref}       % hyperlinks
\usepackage{url}            % simple URL typesetting
\usepackage{booktabs}       % professional-quality tables
\usepackage{amsfonts}       % blackboard math symbols
\usepackage{nicefrac}       % compact symbols for 1/2, etc.
\usepackage{microtype}      % microtypography
\usepackage{lipsum}
\usepackage{caption} % for \captionof
\usepackage{fancyhdr}       % header
\usepackage{graphicx}       % graphics
\graphicspath{{media/}}     % organize your images and other figures under media/ folder

\usepackage{amsmath}
\usepackage[ruled,vlined]{algorithm2e}
\usepackage{array}
\usepackage{float}
\usepackage{makecell}    % for \makecell (multi-line cells)
\usepackage{placeins}    % for \FloatBarrier
\usepackage{multirow}

\title{Continual Learning for Image Captioning through Improved Image-Text Alignment
}
\author{
  Bertram Taetz$^*$\\
  IT \& Engineering \\
  International University of Applied Sciences \\
  Erfurt, Germany\\
  \texttt{Bertram.Taetz@iu.org} \\
   \And
  Gal Bordelius\thanks{These authors contributed equally to this work.} \\
  IT \& Engineering \\
  International University of Applied Sciences \\
  Erfurt, Germany\\
  \texttt{Gal.Bordelius@iu-study.org} \\
}

\begin{document}
\maketitle

\begin{abstract}
Generating accurate and coherent image captions in a continual learning setting remains a major challenge due to catastrophic forgetting and the difficulty of aligning evolving visual concepts with language over time. In this work, we propose a novel multi-loss framework for continual image captioning that integrates semantic guidance through prompt-based continual learning and contrastive alignment. Built upon a pretrained ViT-GPT-2 backbone, our approach combines standard cross-entropy loss with three additional components: (1) a prompt-based cosine similarity loss that aligns image embeddings with synthetically constructed prompts encoding objects, attributes, and actions; (2) a CLIP-style loss that promotes alignment between image embeddings and target caption embedding; and (3) a language-guided contrastive loss that employs a triplet loss to enhance class-level discriminability between tasks. Notably, our approach introduces no additional overhead at inference time and requires no prompts during caption generation. We find that this approach mitigates catastrophic forgetting, while achieving better semantic caption alignment compared to state-of-the-art methods. The code can be found via the following link \url{https://github.com/Gepardius/Taetz_Bordelius_Continual_ImageCaptioning}.
\end{abstract}

% keywords can be removed
\keywords{Image Captioning \and Continual Learning \and Image-Text Alignment}

\begin{minipage}{\textwidth}
\centering
\includegraphics[width=0.9\linewidth,height=0.27\textheight,keepaspectratio]{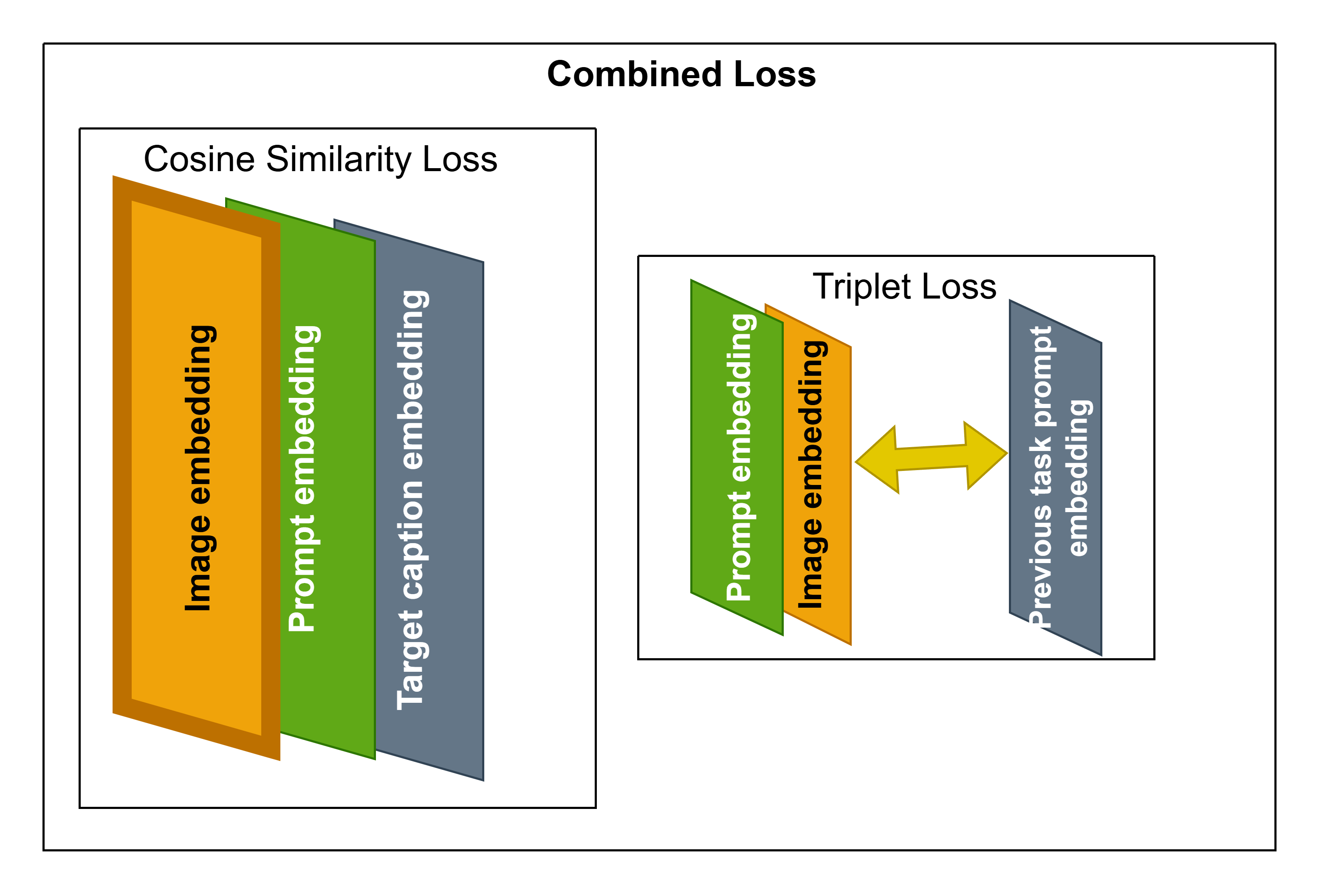}
\captionof{figure}{Overview of the proposed multi-objective training approach combining prompt-based, CLIP-based cosine similarity loss and triplet loss.}
\label{fig:novel-approach}
\end{minipage}

%%%%%%%%% BODY TEXT
\section{Introduction}

Understanding how to generate natural language descriptions of images is a fundamental problem at the intersection of computer vision and natural language processing. Recent advances in deep learning have enabled remarkable progress in static image captioning~\cite{xu2015show}, where large-scale vision-language models are trained to describe images with fluent, contextually appropriate sentences. However, these conventional systems are designed for fixed, closed-world scenarios in which all training data are available upfront, and the models are not required to adapt to new visual categories or linguistic concepts over time. In contrast, \textbf{continual learning} is an emerging research paradigm that seeks to endow captioning systems with the ability to learn from a stream of data composed of \emph{sequentially arriving tasks}, \emph{new domains}, or \emph{novel object categories}, all without revisiting previous data~\cite{parisi2019continual,wang_comprehensive_2024}. This setting introduces unique challenges associated with lifelong learning, such as the risk of \emph{catastrophic forgetting}~\cite{mccloskey_catastrophic_1989}, in which previously acquired knowledge is overwritten as the model adapts to novel information. This paradigm can be applied to the case of image captioning,  \cite{chiaro2020ratt,nguyen_contcap_2020}. Successful continual image captioning systems would be capable of incrementally expanding their understanding and generating accurate, coherent captions for both seen and unseen concepts as they are exposed to them. The motivation for continual image captioning extends beyond academic interest. It is highly relevant for \emph{real-world applications} such as autonomous agents, assistive devices, and interactive robotics, which operate in dynamic environments and must continually assimilate new visual and linguistic patterns~\cite{golroudbari_recent_2023}. Unlike classification, captioning is a generative task that demands \textbf{cross-modal reasoning}\textemdash grounding visual information in language with both fine-grained perceptual detail and pragmatic contextualization. As such, continual image captioning provides a \emph{rich testbed} for studying not only incremental learning and memory-efficient architectures, but also the complex interplay between vision and language adaptation~\cite{hessel2021clipscore}. It invites exploration of new training strategies, self-supervised learning, and regularization schemes specifically tailored for vision-language models in an open world. Despite growing interest, continual learning methods for computer vision have been largely explored in the context of image classification~\cite{wang_comprehensive_2024}, with comparatively less attention to more complex vision-language generation settings. Prior attempts to mitigate catastrophic forgetting using prompt-based methods or contrastive learning have shown preliminary promise in related tasks~\cite{Wang2022, wang2022dualprompt}, but unique challenges arise for generative modeling: the need for \emph{preserving object-word alignments}, \emph{maintaining linguistic diversity}, and \emph{avoiding loss of semantic granularity} across tasks. In this work, we address these challenges by introducing a novel \textbf{multi-loss training framework} for continual image captioning. Building on a pretrained ViT-GPT-2 architecture~\cite{shreydan2023visiongpt2}%\cite{xin_image_nodate}
, our approach combines the standard cross-entropy loss with (1) a \emph{prompt-based cosine similarity loss} ($L_{\text{nouns}}$) to align image embeddings with semantic prompt representations, (2) a \emph{CLIP-based cosine similarity loss} ($L_{\text{CLIP}}$) to align image embeddings to target caption textual embedding. The framework is initiated with prompt-based, later switching to caption-based alignment training. (3) A \emph{language-guided contrastive loss on the class level} (\(L_{\mathrm{LGCL}}\)), similar to~\cite{khan2023introducinglanguageguidancepromptbased}, which promotes semantic discriminability across tasks through a triplet-based cosine similarity scheme. Our proposed \emph{dynamic loss balancing} mechanism further prevents any single loss component from dominating training, reducing the risk of overfitting or under-adaptation. All four losses are summed together into a single loss, which is used for backpropagation. Through experiments on a continual split of the MS-COCO dataset~\cite{nguyen_contcap_2020}, we show that our method outperforms baseline and state-of-the-art approaches, in particular exhibiting improved semantic retention. Note that our method incurs no inference-time overhead, making it attractive for practitioners in resource-constrained environments.
Our contributions can be summarized as follows.
\begin{itemize}
\item We propose a new prompt-based training approach for continual image captioning that reduces catastrophic forgetting as compared to state-of-the-art-methods, by employing a novel composite objective to align image and text embeddings in a continual learning setting.
\item We release code and standardized dataset splits for the two continual MS-COCO benchmarks used in our experiments, enabling objective and reproducible comparisons.
\end{itemize}

% \noindent
% Our contributions are:
% \begin{itemize}
%     %\item The introduce a new continual image captioning framework, evaluated on two MS-COCO continual benchmarks.
%     \item A novel prompt-based approach that uses a multiple loss system with (\(L_{\text{nouns}}\)) and (\(L_{\text{CLIP}}\)) cosine similarity losses to align image embedding with textual embeddings in continuous training, as well as implementing a loss of language guidance based on \cite{khan2023introducinglanguageguidancepromptbased}.
%     %\item Our approach is computationally efficient and does not require additional inference resources compared to static models.
%     \item Code and dataset splits for both MS-COCO continual datasets used for objective comparisons to our method.
% \end{itemize}

\section{Related Work}

Image captioning, the task of generating descriptive natural language sentences for images, has seen remarkable advances through deep learning.
%, most notably via encoder-decoder architectures utilizing convolutional neural networks (CNNs) and recurrent neural networks (RNNs) or transformers \cite{jia_guiding_2015}, \cite{pu_variational_2016}, \cite{vinyals_show_2015}. 
Early approaches, such as the Neural Image Caption Generator (NIC) \cite{vinyals_show_2015}, demonstrated promising results and inspired a series of works aimed at improving accuracy and expressiveness \cite{cornia_show_2019}. Despite progress, these models are typically trained offline on fixed datasets and crucially suffer from catastrophic forgetting when new tasks are encountered sequentially, a phenomenon well documented in the continual learning literature \cite{mccloskey_catastrophic_1989}, \cite{wang_comprehensive_2024}. To address catastrophic forgetting, continual learning methods have emerged and can be largely classified into the following categories of approaches: regularization-based, replay-based, optimization-based, representation-based, architecture-based \cite{wang_comprehensive_2024}. 
%Regularization-based approaches constrain updates to critical model parameters but generally underperform on complex vision-language tasks \cite{mai_online_2021}, \cite{rebuffi_icarl_2017}. Rehearsal-based methods use data replay buffers to mitigate forgetting, but face scalability and privacy concerns \cite{chaudhry_efficient_2019}, \cite{chaudhry_tiny_2019}, \cite{hayes_memory_2019}, \cite{shokri_privacy-preserving_2015}, \cite{lopez-paz_gradient_2022}, \cite{tasar_incremental_2019}. Architecture-based strategies allocate sub-networks or modules to tasks but require increased parameters and often task identity at inference, hindering task-agnostic deployment \cite{rusu_progressive_2022}, \cite{yoon_lifelong_2018}, \cite{li_learn_2019}, \cite{loo_generalized_2020}. A recent survey about continual learning approaches can be found in \cite{wang_comprehensive_2024}.
Prompt-based continual learning is a recent approach that belongs to the category of representation-based approaches and introduces parameter-efficient solutions by leveraging pre-trained models and learnable prompts, synthesizing transfer learning ideas from natural language processing (NLP) and extending them to vision-language models \cite{Wang2022}, \cite{wang2022dualprompt}, \cite{khan2023introducinglanguageguidancepromptbased}.
%\cite{lester_power_2021}, \cite{li_prefix-tuning_2021}. 
This approach bypasses the need to store or replay task data and can encode task information directly in prompts without requiring explicit task IDs \cite{pfeiffer_adapterfusion_2021}. %, \cite{wang_k-adapter_2020}. 
Recent work integrating language guidance into vision models has demonstrated improved generalization, especially for unseen classes, by aligning visual and semantic spaces \cite{radford_learning_2021}, \cite{naeem_i2mvformer_2022}, \cite{naeem_learning_2021}. Within image captioning, the challenge of continual learning is rarely studied. Methods such as ContCap \cite{nguyen_contcap_2020} employed freezing, pseudo-labeling, and feature distillation. The approaches \cite{chiaro2020ratt}, \cite{li_incremental_2024} focussed particularly on recurrent approaches. Yet prompt-based approaches in image captioning remain largely unexplored. Our research advances this frontier by integrating prompt-based continual learning and language guidance, establishing a novel, efficient paradigm for incremental image captioning that addresses catastrophic forgetting without rehearsal or extensive retraining.
\section{Method}

\subsection{Notation}
Continual Learning focuses on training a machine learning model on a data stream originating from a sequence of tasks. Let us denote this sequence of tasks as $\mathcal{D} = \{D_1, D_2, \ldots, D_T\}$, where each task $D_t$ consists of tuples $ (x_{tk}, y_{tk})$. Here, $x_{tk} \in \mathcal{X}$ represents the input image $k$ in task $t$ in the RGB color space, and $y_{tk}  \in \mathcal{Y}$ denotes the corresponding label associated with task $t$.
%Consistent with prior works, the tasks are assumed to be non-overlapping; that is, any given image and its label appear only in a single task, and the model does not retain access to training data from previous tasks once a new task begins. Our focus is on the class-incremental learning scenario, where the identity of the task is not provided at test time.
% Furthermore, we assume the availability of a pre-trained feature extractor $F$ for images, which remains frozen throughout training, as established in previous studies, e.g. \cite{Wang2022}. Similarly, a pre-trained feature extractor for language data is also assumed to be available and similarly kept frozen during training \cite{radford21a}. To ensure consistency with previous literature, the feature encoders for vision and language are trained independently of each other.
For the vocabulary we utilize the pretrained GPT-2 tokenizer \cite{radford2019language}, with a vocabulary of $50,257$ tokens. This vocabulary, originally trained on a WebText corpus, allows for efficient and consistent tokenization of textual inputs without modifying the underlying language model. By leveraging this fixed vocabulary, our method ensures compatibility with the GPT-2 architecture.
\subsection{Proposed Approach}
\begin{figure}[h]
    \centering
    \includegraphics[width=0.9\textwidth]{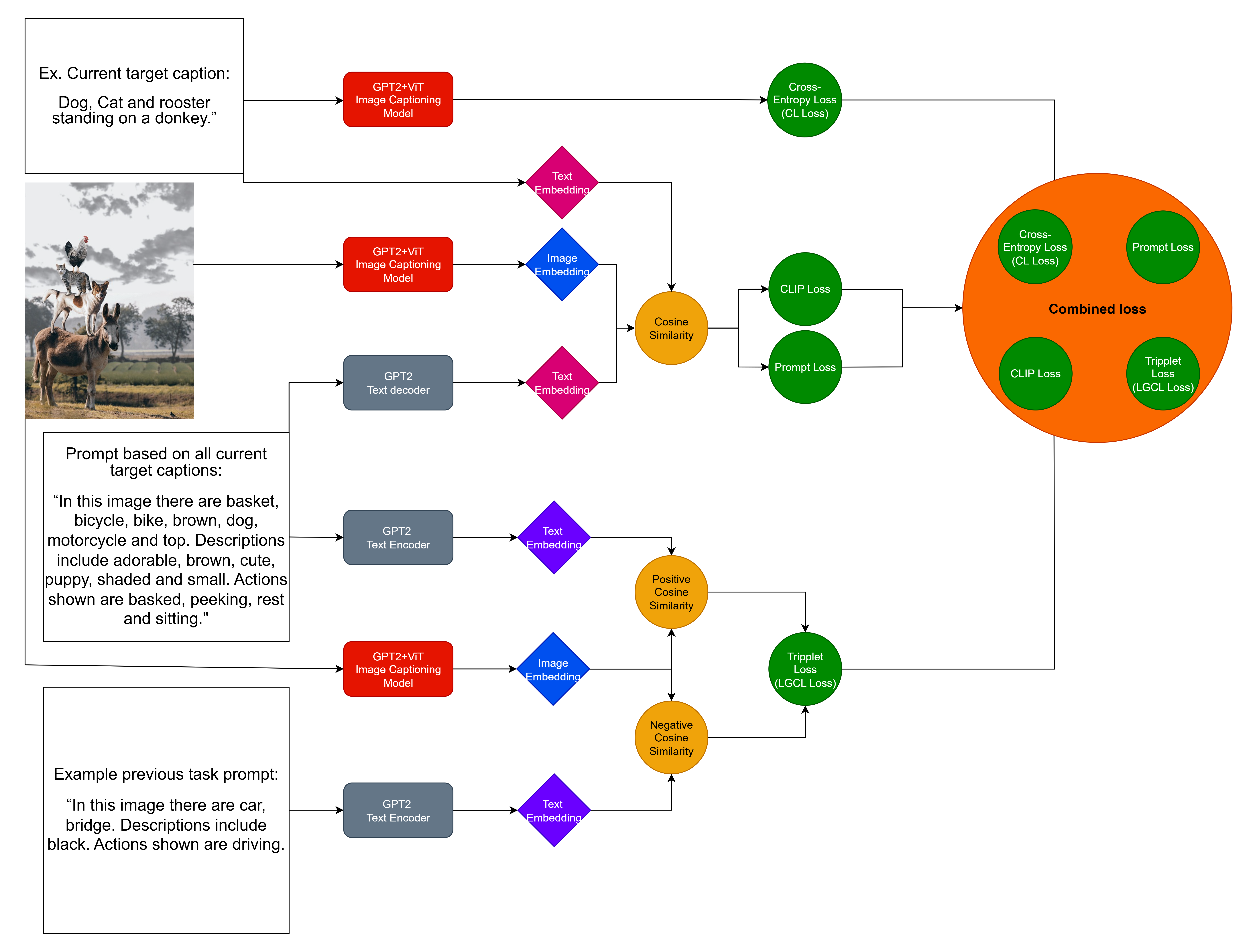}
    \caption{Overview of the proposed multi-objective training approach for image captioning. The model is optimized with four supervisory signals: Standard Cross-Entropy Loss, Prompt based Cosine Similarity Loss, CLIP based Cosine Similarity Loss and Language-Guided Contrastive Loss.}
    \label{fig:novel-approach1}
\end{figure}

% \bigskip
To address the challenge of producing image captions that are both semantically rich and visually discriminative and effective against catastrophic forgetting, we introduce a training framework that integrates multiple supervisory signals. The proposed approach, illustrated schematically in Figure~\ref{fig:novel-approach1}, 

\FloatBarrier
\textbf{Standard Cross-Entropy Loss (\(L_{\mathrm{CE}}\))}\\
% The foundational objective is the standard cross-entropy loss, which serves as the principal source of supervision for the sequence modeling task. For each training instance, given a reference caption \(Y = (y_1, \ldots, y_T)\) with vocabulary size \(V\), and the predicted distribution over tokens \(\hat{Y} = (\hat{y}_1, \ldots, \hat{y}_T)\), the 
The cross-entropy loss is defined as:
\[
L_{\mathrm{CE}}(\theta) = -\sum_{t=1}^{T} \sum_{i=1}^{V} y_{t}^{i} \log \hat{y}_{t}^{i},
\]
where $T$ is the number of words in the ground truth caption, $V$ is the vocabulary size, \(y_{t}^{i}\) is the binary indicator (one-hot encoding) of the true target at position \(t\), \(\hat{y}_{t}^{i} = p_\theta(y_t = i| y_{1:t-1}, I)\) is the model-predicted probability for the \(i\)-th vocabulary term at position \(t\), given the image $I$, the model weights \(\theta\) and the previous words (\(y_{1:t-1}\)). This loss encourages the model to generate linguistically coherent and contextually appropriate captions from image encodings. \\

\textbf{Prompt-Based Cosine Similarity Loss (\(L_{\mathrm{nouns}}\))}\\
To explicitly reinforce the model’s sensitivity to the key visual elements within a scene, we employ a cosine similarity loss between the image representation and a noun, adjective and action centric prompt embeddings. For a given reference caption \(Y\), we extract a subset of tokens corresponding to salient visual entities such as objects (nouns), attributes (adjectives), and actions (verbs). These are arranged into textual prompts, following the GPT-2 format, which is then encoded by a frozen GPT-2 model. Denoting the normalized image and prompt embeddings as \(\mathbf{e(\theta)}_{\text{img}}\) and \(\mathbf{e}_{\text{prompt}}\) respectively. Based on the cosine similarity:
\[
\cos(\mathbf{u}, \mathbf{v}) = \frac{\mathbf{u}^\top \mathbf{v}}{\|\mathbf{u}\| \cdot \|\mathbf{v}\|}.
\]
We define the loss as
\[
L_{\mathrm{nouns}}(\theta) = 1 - \cos\left(\mathbf{e(\theta)}_{\text{img}}, \mathbf{e}_{\text{prompt}}\right).
\]
This formulation ensures that the visual encoder produces embeddings that align semantically with linguistically grounded representations of key entities in the scene. It acts as an auxiliary signal in early training (for our experiments (\(L_{\mathrm{nouns}}\)) was first calculated for first 2 epochs) to improve concept grounding and feature localization.

\textbf{CLIP-Based Cosine Similarity Loss (\(L_{\mathrm{CLIP}}\))}  
In later training epochs, we transition from structured prompts to using the model's own target decoded captions for alignment through a Cosine Similarity loss between the image embedding and the embedding of the target decoded captions. Denoting the normalized image and target caption embeddings as \(\mathbf{e}_{\text{img}}\) and \(\mathbf{e}_{\text{caption}}\) respectively: 

\[
L_{\mathrm{CLIP}}(\theta) = 1 - \cos\left(\mathbf{e(\theta)}_{\text{img}}, \mathbf{e}_{\text{caption}}\right).
\]

This loss encourages alignment between visual and textual modalities in a shared semantic space, reinforcing multimodal consistency as the model becomes more capable of producing fluent and context-aware descriptions.

\textbf{Language-Guided Contrastive Loss (\(L_{\mathrm{LGCL}}\))}
Following recent advances in language guidance for prompt-based continual learning~\cite{khan2023introducinglanguageguidancepromptbased}, we further enhance the model's discriminative capacity by employing a language-guided contrastive objective. Here, we draw inspiration from triplet loss formulations~\cite{feragen2015}, aiming to maximize the alignment between semantically corresponding image-text pairs while repelling mismatching associations.

Let \(\mathbf{v(\theta)}_{\text{img}}\) be the embedding of the image, \(\mathbf{v}_{\text{text}}^{+}\) the embedding of the positive prompt or caption (derived from the current task/noun prompt), and \(\mathbf{v}_{\text{text}}^{-}\) the embedding of a negative prompt (e.g., an unrelated caption from a previous task). 
The triplet contrastive loss is then expressed as:
\[
L_{\mathrm{LGCL}}(\theta) = 1 - \cos(\mathbf{v(\theta)}_{\text{img}}, \mathbf{v}_{\text{text}}^{+}) + \cos(\mathbf{v(\theta)}_{\text{img}}, \mathbf{v}_{\text{text}}^{-}).
\]
This objective encourages \(\mathbf{v(\theta)}_{\text{img}}\) to remain proximate to its correct linguistic counterpart, while maintaining a margin from negatives. % and is based on \cite{khan2023introducinglanguageguidancepromptbased}.
The pseudocode for the algorithm can be found in the appendix \ref{sec:appendix}.

% \textbf{Dynamic Loss Weighting and Total Objective}
% To ensure balanced optimization and avoid any single loss component dominating training dynamics, we employ an adaptive weighting scheme for LGCL loss. Let \(\beta\) denote the normalized contribution of the contrastive loss, computed at each iteration as:
% \[
% \beta = \frac{L_{\mathrm{LGCL}}}{L_{\mathrm{CE}} + L_{\mathrm{nouns}} + L_{\mathrm{CLIP}} + L_{\mathrm{LGCL}}}
% \]

% The total loss used for parameter updates is thus:
% \[
% L_{\mathrm{total}} = L_{\mathrm{CE}} + L_{\mathrm{nouns}} + L_{\mathrm{CLIP}} + L_{\mathrm{LGCL}} \cdot  \beta
% \]
% By dynamically adjusting \(\beta\) and weighing LGCL loss, we avoid LGCL loss overpowering other losses and overadjusting the model's parameters.

% \bigskip

% Importantly, the Prompt-Based Cosine Similarity Loss, CLIP-Based Cosine Similarity Loss and the Language-Guided Contrastive Loss are employed only during the training phase. During inference, the model requires only the input image to generate captions, introducing no additional computational complexity or input requirements.

% \textcolor{red}{Changed!} \\ 

\section{Experiments}

\subsection{Metrics}
We evaluated the performance of our models using standard image captioning metrics, including BLEU \cite{papineni_bleu_2001}, ROUGE \cite{lin_rouge_2004}, CIDEr \cite{vedantam_cider_2015}, and METEOR \cite{lavie_meteor_2007}, used in COCO-Caption. Note, METEOR tracks semantic consistency particularly well at the sentence level, because it aligns hypotheses to references using stems, synonyms as well as paraphrase tables and balances precision and recall \cite{DBLP:journals/corr/ChenFLVGDZ15}. We also report CLIPScore \cite{hessel2021clipscore}, which evaluates caption quality by measuring the cosine similarity between the CLIP image and text embeddings and correlates very well with human judgment\cite{hessel2021clipscore}. In all cases, higher scores indicate better performance.
%All scores are "higher is better" scores. 
% Unlike the other metrics, CLIPScore does not rely on human-written references and instead captures alignment in a shared vision-language embedding space, offering a reference-free estimate of semantic relevance. 
%These metrics all compare the generated captions to human-written reference captions, quantifying their similarity using different linguistic criteria. BLEU measures n-gram precision, evaluating how many sequences of words in the generated caption match the reference captions. ROUGE emphasizes recall, assessing how many overlapping n-grams or longer subsequences are present between the candidate and reference texts. CIDEr weighs n-gram matches by their consensus across multiple references, promoting captions that are relevant and similar to common human annotations. METEOR considers both precision and recall, incorporating stemming and synonym matching to better capture semantic similarity between predicted and reference captions. 
\noindent
To quantify forgetting, we consider an end-of-task forgetting metric and compute the relative change in metric performance for each task as follows:
\[
\resizebox{0.6\textwidth}{!}{
  $\text{Forgetting}~(\%) = 
  \frac{
  \text{Metric after all tasks are trained} - \text{Metric after task was first trained}
  }{
  \text{Metric after task was first trained}
  }$
  }
\]
This expresses the proportion of performance lost due to forgetting.

\subsection{Experimental Setup}
All experiments were performed on a system equipped with an NVIDIA GeForce RTX 4060 Ti GPU. For both ContCap and RATT dataset splits, the setup remained the same. We trained the model for $5$ epochs, with the batch size of 32 and a learning rate (LR) of 1e-5. We used the AdamW optimizer\cite{loshchilov2019decoupledweightdecayregularization}, with default momentum parameters ($\beta_1=0.9$, $\beta_2=0.999$) and a base learning rate of $\eta_0 = \text{LR}/25$, where $\text{LR}$ is the peak learning rate.

\subsection{Comparison to State-of-the-Art Methods}

Table~\ref{tab:comparison_with_contcap}, in the same tabular format as in the original paper \cite{nguyen_contcap_2020} (Table III), shows that our CLICITA model achieves substantial improvements over the best ContCap baselines ($S_{19}$ and $S_{multiple}$) in nearly all key evaluation metrics. The most notable gains are observed in CIDEr, METEOR, and BLEU-4, indicating that CLICITA significantly improves both the relevance and the fluency of generated captions. While ROUGE-L lags slightly behind the best ContCap variant.

\FloatBarrier
\begin{table}[htbp]
  \centering
  \caption{Comparison of Best ContCap ($S_{19}$ and $S_{multiple}$) vs. CLICITA}
  \label{tab:comparison_with_contcap}
  \resizebox{\textwidth}{!}{%
    \begin{tabular}{|l|c|c|c|c|c|}
      \hline
      \textbf{Metric} & \textbf{Best in ContCap ($S_{19}$)} & \textbf{Best in ContCap ($S_{multiple}$)} & \textbf{CLICITA} & \textbf{Improvement ($S_{19}$)} & \textbf{Improvement ($S_{multiple}$)} \\
      \hline
      BLEU-1   & 47.1 ($FD$)  & 53.6 ($FD$)  & \textbf{67.10} & +20.00 & +13.50 \\
      BLEU-4   & 6.6 ($P$)    & 10.5 ($P$)   & \textbf{22.00} & +15.40 & +11.50 \\
      ROUGE-L  & 34.0 ($FD$)  & \textbf{40.0 ($FD$)}  & 36.10 & +2.10  & -3.90 \\
      METEOR   & 11.2 ($D_F$) & 14.5 ($E_F$) & \textbf{27.90} & +16.70 & +13.40 \\
      CIDEr    & 10.0 ($P$)   & 19.2 ($E_F$) & \textbf{66.20} & +56.20 & +47.00 \\
      \hline
    \end{tabular}
  }
\end{table}

% \FloatBarrier

Table~\ref{tab:comparison_ratt_vs_clicita} presents a task-wise comparison on the RATT split, in similar tabular format as in the \cite{chiaro2020ratt}, (Table 2). While CLICITA reports lower BLEU-4 and CIDEr scores across all tasks, indicating a slight drop in n-gram precision and content specificity, it consistently outperforms RATT on METEOR, a key semantic metric. This suggests that CLICITA produces captions with better overall meaning alignment and linguistic fluency, even when exact n-gram matches are fewer. 
%The METEOR improvements are particularly large in tasks like Food, Interior, and Animals, highlighting the model’s strength in semantic generalization across diverse domains.

%\FloatBarrier
\begin{table}[h]
  \centering
  \caption{Comparison of RATT vs. Our Method (CLICITA) across tasks. All values reported after training on the last task.}
  \label{tab:comparison_ratt_vs_clicita}
  \scriptsize % smaller font
  \renewcommand{\arraystretch}{1.1} % compact rows
  \resizebox{\textwidth}{!}{%
  \begin{tabular}{|l|ccc|ccc|ccc|ccc|ccc|}
    \hline
    \multirow{2}{*}{Metric} 
      & \multicolumn{3}{c|}{Transport} 
      & \multicolumn{3}{c|}{Animals} 
      & \multicolumn{3}{c|}{Sports} 
      & \multicolumn{3}{c|}{Food} 
      & \multicolumn{3}{c|}{Interior} \\
    \cline{2-16}
     & RATT & CLICITA & $\Delta$ 
     & RATT & CLICITA & $\Delta$ 
     & RATT & CLICITA & $\Delta$ 
     & RATT & CLICITA & $\Delta$ 
     & RATT & CLICITA & $\Delta$ \\
    \hline
    BLEU-4  & 21.26 & 17.50 & -3.76 
            & 24.68 & 19.70 & -4.98 
            & 31.61 & 19.90 & -11.71 
            & 21.69 & 18.50 & -3.19 
            & 27.27 & 19.80 & -7.47 \\
    METEOR  & 21.69 & 26.50 & +4.81 
            & 23.49 & 30.30 & +6.81 
            & 27.07 & 28.90 & +1.83  
            & 21.10 & 28.40 & +7.30 
            & 22.57 & 29.90 & +7.33 \\
    CIDEr   & 63.49 & 58.30 & -5.19 
            & 72.49 & 64.30 & -8.19 
            & 80.85 & 56.60 & -24.25 
            & 51.95 & 50.60 & -1.35 
            & 65.36 & 51.50 & -13.86 \\
    \hline
  \end{tabular}}
\end{table}
In the following sections we evaluated our method on two different datasets in  two distinct settings to investigate the improvement of the designed mechanism over the baseline pretrained model. Note that pre-trained models already serve the continual learning setting due to flat minima more often found in pre-trained models \cite{wang_comprehensive_2024}. The models are:
\begin{itemize}
     \item \textbf{CLICITA}: Combines all loss functions ($L_{\text{CE}}$, $L_{\text{nouns}}$, $L_{\text{CLIP}}$ and $L_{\text{LGCL}}$).
     \item \textbf{Pre-Trained Basemodel}: This is a baseline pre-trained image-captioning model without continual-learning mechanisms.
 \end{itemize}

\subsection{Experimental Result - ContCap Dataset Split}
Following the training approach outlined in ContCap \cite{nguyen_contcap_2020}, we incrementally introduced five object classes. These five classes (tasks) were sequentially added one by one into our model: person, sports ball, tv, toilet, and bottle, with the class 'bottle' being the final task trained for the model.
The focus is on knowledge retention. Table \ref{tab:avg_forgetting_comparison_ContCap} shows the average forgetting scores for this dataset. 
\begin{table}[h]
\centering
\caption{Comparison of average total forgetting across ContCap and RATT Splits. Lower values indicate better knowledge retention. Bold highlights indicate best results per metric.}
\resizebox{\textwidth}{!}{
\begin{tabular}{|l|c|c|c|c|c|c|c|}
\hline
\textbf{Split} & \textbf{BLEU-1} & \textbf{BLEU-4} & \textbf{ROUGE-L} & \textbf{METEOR} & \textbf{CIDEr} & \textbf{CLIP} & \textbf{Average Forgetting} \\
\hline
\multicolumn{8}{|c|}{\textit{Average Forgetting (\%)}} \\
\hline
CLICITA (ContCap)  & \textbf{-0.32} & \textbf{-2.42} & \textbf{-0.96} & 2.66  & -4.42 & -0.44 & \textbf{-5.9} \\
Pre-Trained Basemodel (ContCap) & -1.18          & -3.10          & -1.68          & \textbf{3.72}           & \textbf{-4.06}          & \textbf{-0.36}          & -6.66 \\
\hline
% CLICITA (RATT)     & \textbf{-0.14} & \textbf{-2.50} & \textbf{-1.36} & \textbf{0.24}  & \textbf{-3.62} & -0.68 & \textbf{-7.06} \\
% Baseline (RATT)    & -0.50          & -3.60          & -2.00          & -0.76          & -3.76          & \textbf{-0.60}          & -11.22 \\
% \hline
\end{tabular}
}
\label{tab:avg_forgetting_comparison_ContCap}
\end{table}
\noindent
As can be observed from Table~\ref{tab:avg_forgetting_comparison_ContCap}, our CLICITA method achieves overall better knowledge retention than Pre-Trained Basemodel.

\subsection{Experimental Results - RATT dataset split}
% some parts of creating their dataset split are ambiguous, with randomly selecting 50\% of val images for testing, and when we tried to split into same tasks, 
 
% We tried to replicate how RATT split their dataset, however we received more images per task, therefore we also had to limit to the exact number of images that they used in order to have somewhat comparable results. 
In RATT \cite{chiaro2020ratt} split they define five distinct tasks (Transport, Animals, Sports, Food, Interior) based on object categories and processes the dataset to ensure non-overlapping image assignments across tasks. The implementation strictly enforces the paper's reported image counts per task (14,266/3,431/3,431 for Transport, 9,314/2,273/2,273 for Animals, etc.) by first filtering images containing relevant categories, removing duplicates across tasks, and maintaining only images with more or equal to 5 captions and keeping only the first 5 captions. The validation set is further split 50/50 into validation and test sets. 
We compare our \textbf{CLICITA method} against Pre-Trained Basemodel, again with a focus is on knowledge retention.

\begin{table}[h]
\centering
\caption{Comparison of Average and Total Forgetting Across ContCap and RATT Splits. Lower values indicate better knowledge retention. Bold highlights indicate best results per metric.}
\resizebox{\textwidth}{!}{
\begin{tabular}{|l|c|c|c|c|c|c|c|}
\hline
\textbf{Split} & \textbf{BLEU-1} & \textbf{BLEU-4} & \textbf{ROUGE-L} & \textbf{METEOR} & \textbf{CIDEr} & \textbf{CLIP} & \textbf{Average Forgetting} \\
\hline
\multicolumn{8}{|c|}{\textit{Average Forgetting (\%)}} \\
% \hline
% CLICITA (ContCap)  & \textbf{-0.32} & \textbf{-2.42} & \textbf{-0.96} & 2.66  & -4.42 & -0.44 & \textbf{-5.9} \\
% Baseline (ContCap) & -1.18          & -3.10          & -1.68          & \textbf{3.72}           & \textbf{-4.06}          & \textbf{-0.36}          & -6.66 \\
\hline
CLICITA (RATT)     & \textbf{-0.14} & \textbf{-2.50} & \textbf{-1.36} & \textbf{0.24}  & \textbf{-3.62} & -0.68 & \textbf{-7.06} \\
Pre-Trained Basemodel (RATT)    & -0.50          & -3.60          & -2.00          & -0.76          & -3.76          & \textbf{-0.60}          & -11.22 \\
\hline
\end{tabular}
}
\label{tab:avg_forgetting_comparison_RATT}
\end{table}

Again, as can be observed in Table~\ref{tab:avg_forgetting_comparison_RATT}, 
our CLICITA method achieves better knowledge retention as compared to Pre-Trained Basemodel. In TAble~\ref{res:qualitytive_CLICITA} shows qualitative results of the proposed method after different training stages for different tasks. It can be observed that the captioning is reasonable and semantically similar to the target.

\FloatBarrier
\begin{table}[H]
\centering
\caption{Qualitative results on RATT dataset split of our CLICITA model}
\begin{tabular}{
  >{\centering\arraybackslash}p{9cm}
  >{\centering\arraybackslash}p{3cm}
  >{\centering\arraybackslash}p{3cm}}
\toprule
\textbf{Task / Target Caption} & \textbf{After training each task} & \textbf{After training task 5 (Interior)} \\
\midrule

\makecell{
  \textbf{Task 1: Transport}\\
  Target:\\
  \emph{``A passenger bus that is driving down the street''}\\[4pt]
  \includegraphics[width=3.5cm]{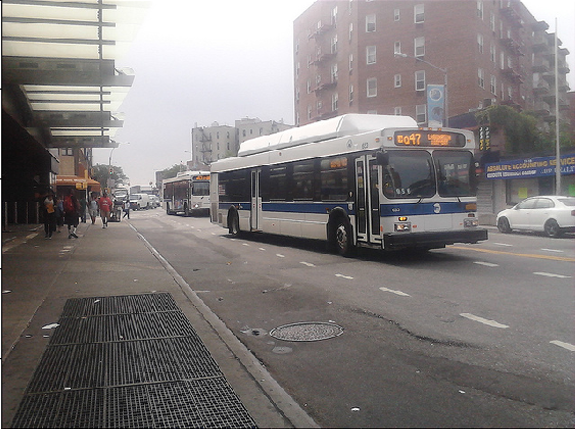}
}
& ``public transit bus on a city street.'' 
& ``public transit bus traveling down a city street.'' \\
\midrule

\makecell{
  \textbf{Task 2: Animals}\\
  Target:\\
  \emph{``A number of zebras standing in the dirt near a wall''}\\[4pt]
  \includegraphics[width=3.5cm]{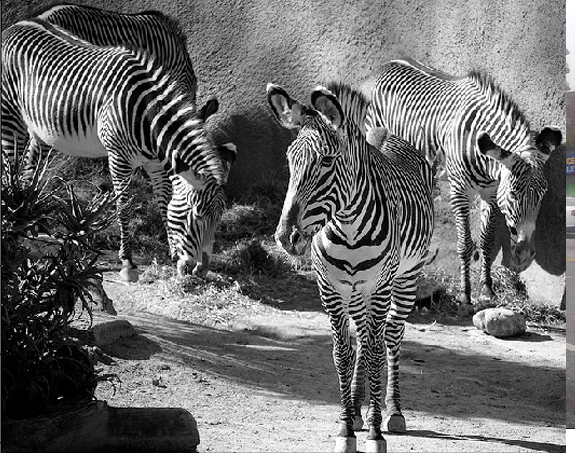}
}
& ``zebra standing next to a group of zebras.'' 
& ``zebra standing next to a bunch of other zebras.'' \\
\midrule

\makecell{
  \textbf{Task 3: Sport}\\
  Target:\\
  \emph{``A man is holding a surfboard and staring out into the ocean''}\\[4pt]
  \includegraphics[width=3.5cm]{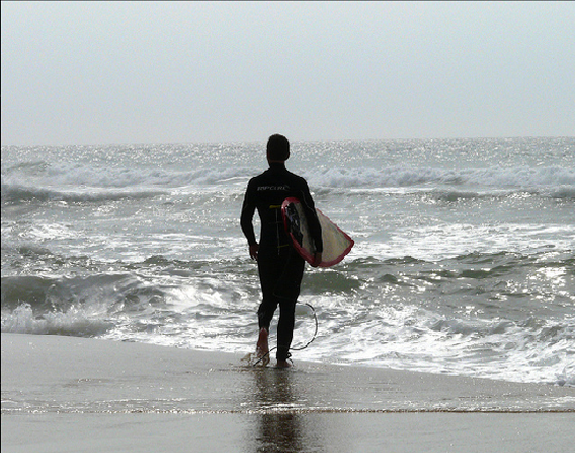}
}
& ``man carrying a surfboard on the beach.'' 
& ``man standing in the water holding a surfboard.'' \\
\midrule

\makecell{
  \textbf{Task 4: Food}\\
  Target:\\
  \emph{``A woman sells cupcakes with fancy decorations on them''}\\[4pt]
  \includegraphics[width=3.5cm]{woman.png}
}
& ``woman is holding a cupcake with a sign on it.'' 
& ``woman is standing in front of a cupcake display.'' \\
\bottomrule
\end{tabular}
\label{res:qualitytive_CLICITA}
\end{table}
\section{Discussion}

% Across both the ContCap and RATT dataset splits, our CLICITA method demonstrates improved knowledge retention, exhibiting consistently lower forgetting compared to the baseline. While there is a slight reduction in overall peak performance metrics. These experimental results highlight the method’s effectiveness in mitigating catastrophic forgetting while preserving competitive captioning quality.
Overall, across both ContCap and RATT dataset splits and as compared against the original state-of-the-art methods, our CLICITA method achieves mostly better knowledge retention, with lower average forgetting, in particular in metrics addressing semantic consistency. Moreover, CLICITA performs overall favorably compared to the pre-trained basemodel, with respect to average forgetting scores of different metrics in the mentioned datasets.\footnote{Note, the scores for each task can be reproduced via the code that will be shared.}
% As mentioned, all experiments were performed on a system equipped with an NVIDIA GeForce RTX 4060 Ti GPU, which imposed practical constraints on both the scale and speed of experimentationfactors directly impacting the training cost. Nonetheless, this experimental setup demonstrates the feasibility of the proposed framework in resource-constrained environments. To further substantiate our findings and assess the generalizability of the approach, future work should involve benchmarking on a broader range of continual image captioning datasets. Such validation would enable a more comprehensive evaluation across diverse visual and linguistic domains, and potentially uncover additional trade-offs.
\section{Conclusion}
In this work, we introduced CLICITA, a novel continual image captioning framework that integrates prompt-based semantic guidance with multiple image-text alignment losses. By combining cross-entropy with prompt-based cosine similarity, CLIP-style cosine similarity loss alignment, and language-guided contrastive learning, our method effectively mitigates catastrophic forgetting while maintaining strong caption generation performance. Notably, our approach introduces no inference-time overhead, making it suitable for deployment in resource-constrained environments. Experimental results on continual MS-COCO benchmarks demonstrate that CLICITA significantly outperforms existing ContCap baseline in most of the image captioning metrics, while outperforming RATT baseline in semantic METEOR metric. Future work can explore the method in stronger and more robust models, as well as on a bigger continual learning image captioning dataset.

%%%%%%%%% BODY TEXT

\section*{Acknowledgments}
The authors have no competing interests to declare that are relevant to the content of this article.
This work was supported by the IU incubator project "Interactive Motion Agent for XR" (IMA-XR), with the project number PR$0311$.
%Bibliography
\bibliographystyle{unsrt}  
\bibliography{references}  

\appendix 
\section{Algorithms}
\label{sec:appendix}

\begin{algorithm}[H]
\caption{Continual Captioning Training with Language-Guided Losses (LGCL, Nouns, CLIP)}
\KwIn{Batch of \texttt{images}, \texttt{input\_ids}, \texttt{labels}, \texttt{prompts}, task\_num, epoch}
\KwOut{Updated model parameters minimizing total loss}
\SetAlgoLined
\DontPrintSemicolon

\SetKwFunction{FEncode}{encode\_text}
\SetKwFunction{FNormalize}{normalize}
\SetKwFunction{FCosSim}{cosine\_similarity}
\SetKwFunction{FDecode}{decode\_captions}

\For{(images, input\_ids, labels, prompts) \textbf{in} train\_dl}{
    $\mathcal{L}_{\text{ce}}, \mathbf{e}_{\text{img}}, \_ \gets$ model(images, input\_ids, labels)\;
    $\mathcal{L}_{\text{lgcl}} \gets 0$,\quad $\mathcal{L}_{\text{nouns}} \gets 0$,\quad $\mathcal{L}_{\text{clip}} \gets 0$\;

    \If{use\_lgcl}{
        prompt\_ids $\gets$ tokenizer(prompts)\;
        $\mathbf{e}_{\text{img}} \gets$ \FNormalize{$\mathbf{e}_{\text{img}}$}\;

        \uIf{epoch $\leq 2$}{
            $\mathbf{e}_{\text{nouns}} \gets$ \FNormalize{\FEncode{prompt\_ids}}\;
            $\mathcal{L}_{\text{nouns}} \gets 1 -$ \FCosSim{$\mathbf{e}_{\text{img}}, \mathbf{e}_{\text{nouns}}$}\;
        }
        \Else{
            decoded $\gets$ \FDecode{labels}\;
            caption\_ids $\gets$ tokenizer(decoded)\;
            $\mathbf{e}_{\text{caption}} \gets$ \FNormalize{\FEncode{caption\_ids}}\;
            $\mathcal{L}_{\text{clip}} \gets 1 -$ \FCosSim{$\mathbf{e}_{\text{img}}, \mathbf{e}_{\text{caption}}$}\;
        }

        \If{task\_num == 0}{
            $\mathbf{e}_{\text{pos}} \gets$ \FNormalize{\FEncode{prompt\_ids}}\;
            \ForEach{$\vec{v}$ \textbf{in} $\mathbf{e}_{\text{pos}}$}{
                append $\vec{v}$ to current\_task\_pool\;
            }
        }

        \If{task\_num > 0}{
            $\mathbf{e}_{\text{pos}} \gets$ \FNormalize{\FEncode{prompt\_ids}}\;
            \ForEach{$\vec{v}$ \textbf{in} $\mathbf{e}_{\text{pos}}$}{
                append $\vec{v}$ to current\_task\_pool\;
            }

            \If{len(neg\_prompt\_pool) $\geq B$}{
                $\mathbf{E}_{\text{neg}} \gets$ \FNormalize{stack(subset of neg\_prompt\_pool)}\;
                $\mathbf{S} \gets \mathbf{e}_{\text{img}} \cdot \mathbf{E}_{\text{neg}}^\top$\;
                $j^* \gets \arg\min_j \mathbf{S}_{ij}$\;
                $\mathbf{e}_{\text{neg}}^{(i)} \gets \mathbf{E}_{\text{neg}}[j^*]$\;
            }
            \Else{
                $\mathbf{e}_{\text{neg}} \gets$ \FNormalize{broadcast(neg\_prompt\_pool[0])}\;
            }

            $s^+ \gets$ \FCosSim{$\mathbf{e}_{\text{img}}, \mathbf{e}_{\text{pos}}$}\;
            $s^- \gets$ \FCosSim{$\mathbf{e}_{\text{img}}, \mathbf{e}_{\text{neg}}$}\;
            $\mathcal{L}_{\text{lgcl}} \gets \text{mean}(\max(0, 1 - s^+ + s^-))$\;
        }
    }
    \textit{// Total loss}\;
    $\mathcal{L}_{\text{total}} \gets \mathcal{L}_{\text{ce}} + \mathcal{L}_{\text{lgcl}} + \mathcal{L}_{\text{nouns}} + \mathcal{L}_{\text{clip}}$\;

    \textit{// Backpropagate and optimize:}\;
    \quad $\mathcal{L}_{\text{total}}.\texttt{backward()}$, optimizer.step(), optimizer.zero\_grad()\;
}
\end{algorithm}

%\FloatBarrier

%\textcolor{red}{No changes} 

\begin{algorithm}[H]
\caption{Inference Algorithm (Caption Generation)} 
\KwIn{Image path, temperature ($t$), deterministic flag}
\KwOut{Generated caption, tokens, logits, loss}
\SetAlgoLined
\DontPrintSemicolon

\SetKwFunction{FGenerate}{generate\_caption}
\SetKwFunction{FDecode}{tokenizer.decode}

% Algorithm body
\textit{// Load and preprocess image}\;
image $\gets$ Image.open(image\_path).convert('RGB')\;
image $\gets$ Apply transforms (resize, normalize, to tensor)\;
image $\gets$ image.unsqueeze(0).to(device)\;

\textit{// Initialize sequence with BOS token}\;
sequence $\gets$ tensor([BOS\_TOKEN\_ID]).to(device)\;

\textit{// Generate caption autoregressively}\;
tokens, logits, loss $\gets$ model.generate(
    image, 
    sequence, 
    max\_tokens=50, 
    temperature=$t$, 
    deterministic=deterministic
)\;

\textit{// Decode tokens to text}\;
caption $\gets$ \FDecode{tokens}\;

\Return{caption, tokens, logits, loss}
\end{algorithm}

\end{document}